\documentclass[conference]{IEEEtran}
\IEEEoverridecommandlockouts

\makeatletter
 \let\old@ps@headings\ps@headings
 \let\old@ps@IEEEtitlepagestyle\ps@IEEEtitlepagestyle
 \def\confheader#1{%
 \def\ps@headings{%
 \old@ps@headings%
 \def\@oddhead{\strut\hfill#1\hfill\strut}%
 \def\@evenhead{\strut\hfill#1\hfill\strut}%
 }%
 \def\ps@IEEEtitlepagestyle{%
 \old@ps@IEEEtitlepagestyle%
 \def\@oddhead{\strut\hfill#1\hfill\strut}%
 \def\@evenhead{\strut\hfill#1\hfill\strut}%
 }%
 \ps@headings%
 }
 \makeatother
\usepackage{tipa}
\usepackage{amsmath,amssymb,amsfonts}
\usepackage[bookmarks=false,hidelinks]{hyperref}
\usepackage{algorithmic}
\usepackage[noadjust]{cite}
\usepackage{graphicx}
\usepackage{textcomp}
\usepackage{multirow}
\usepackage{tabularx}
\usepackage{threeparttable}

\confheader{%
 27\textsuperscript{th} Iranian  Conference on Electrical Engineering (ICEE2019)
 }
 \usepackage[pscoord]{eso-pic}
\newcommand{\placetextbox}[3]{
 \setbox0=\hbox{#3}
 \AddToShipoutPictureFG*{ \put(\LenToUnit{#1\paperwidth},\LenToUnit{#2\paperheight}){\vtop{{\null}\makebox[0pt][c]{#3}}}
 }
 }
 \placetextbox{.23}{0.055}{\small{978-1-7281-1508-5/19/\$31.00~\copyright 2019 IEEE} \cite{8786505}}

\title{Persian Keyphrase Generation \\ Using Sequence-to-sequence Models\\}

\author{

\IEEEauthorblockN{Ehsan Doostmohammadi}
\IEEEauthorblockA{\textit{Graduate Student} \\
\textit{Computational Linguistics Group}\\
\textit{Sharif University of Technology}\\
Tehran, Iran \\
e.doostm72@student.sharif.edu}
\and

\IEEEauthorblockN{Mohammad Hadi Bokaei}
\IEEEauthorblockA{\textit{Assistant Professor} \\
\textit{Information Technology Department}\\
\textit{ICT Research Institute}\\
Tehran, Iran \\
mh.bokaei@itrc.ac.ir}
\and

\IEEEauthorblockN{Hossein Sameti}
\IEEEauthorblockA{\textit{Associate Professor} \\
\textit{Computer Engineering Department}\\
\textit{Sharif University of Technology}\\
Tehran, Iran \\
sameti@sharif.edu}
}

\begin{document}
\maketitle

\begin{abstract}
Keyphrases are a very short summary of an input text and provide the main subjects discussed in the text. Keyphrase extraction is a useful upstream task and can be used in various natural language processing problems, for example, text summarization and information retrieval, to name a few. However, not all the keyphrases are explicitly mentioned in the body of the text. In real-world examples there are always some topics that are discussed implicitly. Extracting such keyphrases requires a generative approach, which is adopted here. In this paper, we try to tackle the problem of keyphrase generation and extraction from news articles using deep sequence-to-sequence models. These models significantly outperform the conventional methods such as Topic Rank, KPMiner, and KEA in the task of keyphrase extraction \footnote{The data and the code can be found in this project's Github repository: \href{https://github.com/edoost/perkey}{https://github.com/edoost/perkey}}.

\end{abstract}


\section{Introduction}
Keyphrases are single words or sequences of words that express the main topics discussed in a piece of text. Knowing the main topics discussed in a text can play an important role in a variety of downstream tasks such as text categorization \cite{hulth2006study}, opinion mining \cite{berend2011opinion}, information retrieval \cite{frank1999domain} and text summarization \cite{zhang2004world}. Keyphrases themselves can be regarded as very dense summarization of the input text which can come of help in many problems.

Keyphrase extraction is quite a well-known task in Natural Language Processing, the purpose of which is to extract words pertaining to the main topics discussed in the input piece of text. The purpose of keyphrase extraction task is to extract explicit keyphrases from a given text. Keyphrase generation on the other hand, is a task of extracting explicit and implicit keyphrases. In real-world examples, there are always keyphrases that are not present in the article, but rather hinted at implicitly. Since we are dealing with implicit keyphrases in this paper, such a task cannot be treated as a sequence labeling, phrase/word classification or scoring problem. In this paper we take an approach of deep sequence-to-sequence learning, similar to neural machine translation.

In the next section we discuss the different approaches to keyphrase extraction and generation in English and Persian. In section III, we define our problem and elaborate on the specific approach we take in this paper. In section IV, we discuss the training and test data, implementation details, baseline models, and our evaluation metrics. Then, we discuss the results in section V and VI. We conclude the paper and discuss the future work in section VII.

\section{Previous Work}
\subsection{Work on English}
Most keyphrase extraction algorithms operate in two steps; they first make a list of keyphrase candidates, and then choose the keyphrases from them \cite{hasan2014automatic}.
The candidate list is made according to several features, namely statistical (e.g. $tf \times idf$), structural (e.g. location of the candidate keyphrase in the text), syntactic (e.g. Part of Speech tags), and external resource features (e.g. Wikipedia titles) \cite{hasan2014automatic}.
As for the second step, the early approaches to keyphrase extraction deal with it as a problem of binary classification \cite{turney1999learning,witten1999kea}; whether a candidate phrase in a text can be classified as 1, present in keyphrases, or 0, not present in the keyphrases, using different approaches such as
decision trees \cite{turney1999learning},
na\"{i}ve Bayes \cite{witten1999kea},
maximum entropy \cite{yih2006finding},
and support vector machines \cite{romary2010automatic}. 
A binary classifier classifies each keyphrase independent of the other keyphrases, hence it may classify several keyphrases pointing to a similar subject. To tackle this drawback, Jiang et al. \cite{jiang2009ranking} offered a pairwise ranking system. 

However, a substantial number of approaches to keyphrase extraction are unsupervised, including graph-based ranking and topic-based clustering. In graph-based ranking it is assumed that keyphrases are the most important words/phrases in a document and if we build a graph in which words/phrases are the nodes and the edges represent the relation between them, we can find the keyphrases by ranking its nodes according to their importance using a graph-based ranking method \cite{brin1998anatomy}.
Topic-based clustering methods on the other hand, try to cluster candidate
keyphrases in a document into topics. For instance KeyCluster \cite{liu2009clustering} clusters  semantically  similar  candidates,  selects  the  candidates  close to  the  centroid  of  each  cluster and Topical PageRank \cite{liu2010automatic} runs TextRank once for each topic induced by LDA.

As for deep approaches to keyphrase extraction, Zhang et al. \cite{zhang2016keyphrase} use joint-layer recurrent neural networks to extract keyphrases from tweets, treating the problem as a sequence labelling task. Another work that takes the same approach is  \cite{zhang2018encoding} that aims to extract keyphrases from microblog posts with the help of encoding the context.

However, in this work we also focus on the problem of keyphrase generation, as well as extraction. The first work to address this problem using deep neural networks is that of Meng et al. \cite{meng2017deep} which tries to tackle the problem using an encoder-decoder, attention, and copying mechanism. Their bidirectional RNN with gated recurrent units cannot outperform non-deep approaches in extracting present keyphrases from most of the datasets, but Copynet \cite{gu2016incorporating} does significantly. Some other works taking the same approach are \cite{zhang2018keyphrase} and \cite{zhang2017deep}, the latter of which uses a convolutional sequence-to-sequence model.

As for the other works, Zhang et al. \cite{zhang2018encoding} try to tackle the problem of keyphrase extraction from microblogs. They make use of the context, which is other blog posts, to inform the model of the previous events. Chen et al. \cite{chen2018keyphrase} address the problem of keyphrase duplication and try to solve it by taking previous phrases into account.

\subsection{Work on Persian}
Literature of keyphrase/keyword extraction from Persian texts is quite minimal. Mohammadi and Analoui \cite{ACCSI13_067}, after removing stop words and stemming, extracts keywords using TF (Term Frequency), TTF (Total Term Frequency), and DF (Document Frequency) matrices and fuzzification.

Khozani and Bayat \cite{khozani2011specialization}, after stemming and removing stop words, uses TFIDF (Term Frequency times Inverse Document Frequency) to calculate the weights of the words in the documents. After calculating the weights of the tokens, the weights of the bi-token keyphrases are calculated using a co-occurrence matrix.

Kian and Zahedi \cite{kian2013improving} uses attention attractive strings
to improve keyword extraction from 800 documents from Hamshahri
news collection \cite{aleahmad2009hamshahri}. Three to seven keyphrases were assigned to each document, the length of each keyphrase ranging from two to four words. The best result, an F\textsubscript{1} of $40.23$, is achieved using attention attractive strings and training on 400 documents.

\section{Methodology}

\begin{figure*}
    \centering
    \includegraphics{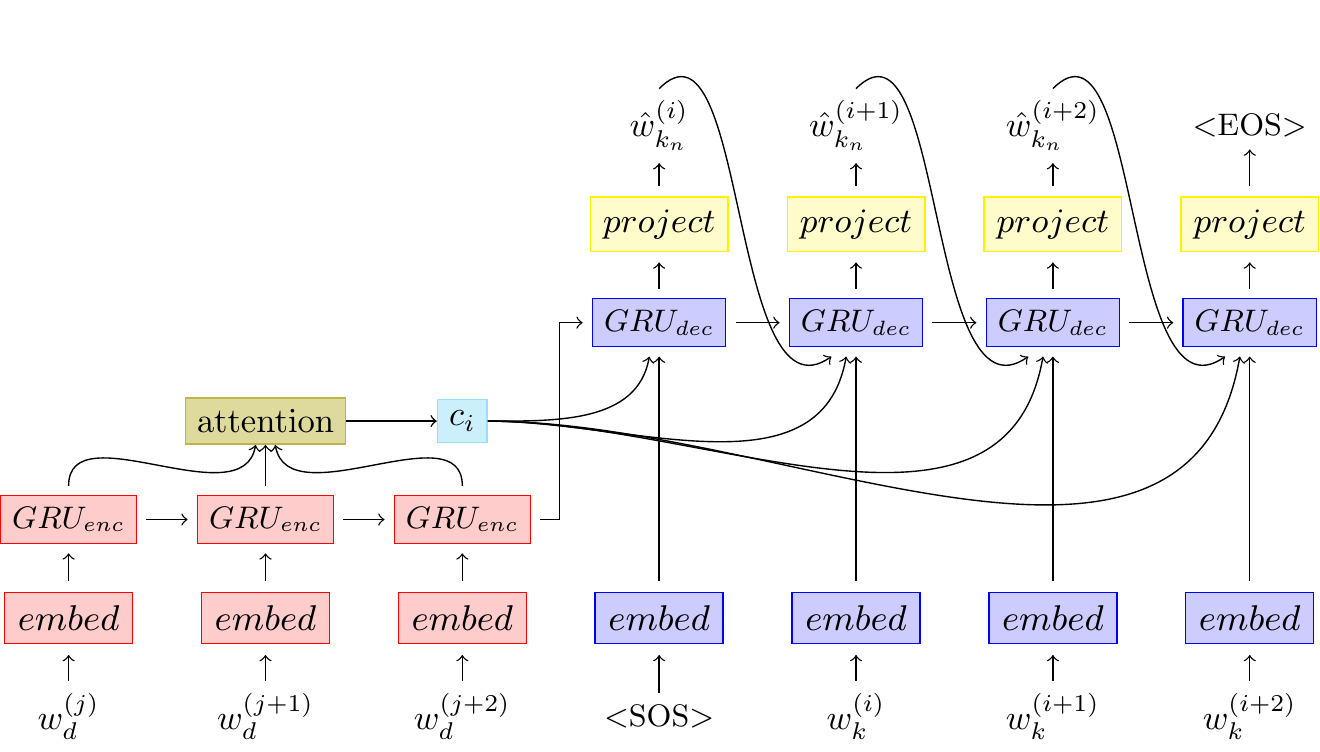}
    \caption{The encoder-decoder model with attention mechanism. In decoder, we feed the gold $w$ in each timestep when we are training and predicted $\hat{w}$ when performing evaluation.}
\end{figure*}

In this paper, we use an encoder-decoder model \cite{cho2014learning,sutskever2014sequence} with Bahdanau's attention mechanism \cite{bahdanau2014neural}. In the training phase, the data is converted from $(x_j, y_{j_i}, y_{j_{i + 1}}, ..., y_{j_T})$ format to $(x_j, y_{j_i}), (x_j, y_{j_{i + 1}}), ..., (x_j, y_{j_T})$, $x$ being the concatenation of the news article's title and body, and $y$, it's keyphrases. It means that a training sample is repeated to the number of its keyphrases which increases training samples from the number of the news articles to the number of their keyphrases.

\subsection{Encoder-Decoder Model}
An encoder-decoder model consists of two RNNs, the last hidden layer or the concatenation of all hidden layers of the first one considered an encoded representation of the input document the second one is going to decode. Here, the input for the encoder is the concatenation of the title and the body of the news and the output of the decoder, the words in the keyphrase. The hidden layer of the encoder, as explained in \cite{cho2014learning}, is calculated as:

\begin{equation}
    h_j = f(x_j, h_{j−1})
\end{equation}

in which, $f$ is a nonlinear function (as we use GRU, $f$ is a Gated Recurrent Unit here) which takes the input $x$ at timestep $j$ and the hidden state of the last timestep, $h_{j-1}$. By obtaining $h_j$, the hidden state of the encoder at timestep $j$, we can easily calculate $p(y_i|y_1, \dots, y_{i−1}, X)$, the probability of producing the output $y$ at timestep $i$, given the previous $y$'s and the input $X$, by the equation below:

\begin{equation}
    s_i = q(s_{i-1}, y_{i-1}, c)
\end{equation}

$s_{i-1}$ being the representation of all previous $y$'s, except the most previous one and
$c$, the context vector, being the representation of $X$, which is either the output of the last timestep, $h_T$, or the context vector computed by the attention mechanism.

\subsection{Attention Mechanism}
Attention mechanism \cite{bahdanau2014neural} gives different weights to different timestemps from the encoder, presuming that different words in an input sentence have different levels of importance to a timestep of the output. So, a hidden state in the decoder is calculated as $s_i = q(s_{i−1}, y_{i−1}, c_i)$. The context vector $c_t$ can be obtained using the formula below:

\begin{equation}
    c_i = \sum_{j=1}^{T_x} \alpha_{ij} h_j
\end{equation}

$\alpha_{ij}$ being the weight of the state $h_j$ of the encoder to the state $s_i$ of the decoder which is calculated as:

\begin{equation}
    \alpha_{ij} = \frac{exp(a(s_{i−1}, h_j))}{\sum_{k=1}^{T_x} exp(a(s_{i−1}, h_k))}
\end{equation}

where $a$ is a feed forward neural network, a soft alignment (not a latent variable as in traditional machine translation), a model which scores how well the inputs around position $j$ and the output at position $i$ match. You can see the schema of the model in figure 1.

\section{Experiments}
\subsection{Training and Testing Datasets}
Here, we use a subset of the PerKey dataset introduced in \cite{doost2018perkey} with at least 3 keyphrases for each news article. As concluded in PerKey paper, news articles with at least 3 keyphrases are more reliable in terms of recall.

The dataset is stored in JSON format, each news article containing the following information:

\noindent
\begin{verbatim}
{title, body, summary, keyphrases, category, url}
\end{verbatim}

The data used in this paper comprises 395,645 news articles crawled from 6 news websites and agencies which provided high-quality keyphrases. After cleaning the data, an assessment was conducted by 5 human evaluators to ensure the quality of the keyphrases. The result was an F\textsubscript{1}-score of 4.264 which guarantees the quality of the keyprhases. As shown in table I, most of the articles contain less than 300 tokens and only 25\% of them contain more than that.

\begin{table}[h]
\caption{Number of Articles Based on Number of Tokens in them}
\label{Table:len_body}
\begin{center}
\begin{tabular}{c|cc}
\# of tokens & \# of articles & \% of total \\
\hline
\hline
40-100 & 72,467 & 18.31\% \\
\hline
100-200 & 129,996 & 32.85\% \\
\hline
200-300 & 92,691 & 23.42\% \\
\hline
300-400 & 59,989 & 15.16\% \\
\hline
400-500 & 40,502 & 10.23\% \\
\hline
total & 395,645 & 100\% \\
\end{tabular}
\end{center}
\end{table}

It appears that half of the articles have 3 keyphrases and 20\% of them have 4. More information on the number of the keyphrases in news articles can be found in table II.

\begin{table}[h]
\caption{Number of Articles Based on their Number of Keyphrases}
\label{Table:len_body}
\begin{center}
\begin{tabular}{c|cc}
\# of keyphrases & \# of articles & \% of total \\
\hline
\hline
3 & 202,748 & 51.24\% \\
\hline
4 & 81,500 & 20.59\% \\
\hline
5 & 56,278 & 14.22\% \\
\hline
6 \& more & 55,119 & 13.8\% \\
\hline
total & 395,645 & 100\% \\
\end{tabular}
\end{center}
\end{table}

Table V shows that most of the keyphrases contain either 1 or 2 tokens. News articles containing keyphrases with more than 7 tokens in them were removed.

\begin{table}[h]
\caption{Number of Keyphrases Based on the Number of Tokens in them}
\label{Table:len_body}
\begin{center}
\begin{tabular}{c|cc}
\# of tokens & \# of keyphrases & \% of total \\
\hline
\hline
1 & 874,685 & 50.65\% \\
\hline
2 & 586,481 & 33.96\% \\
\hline
3 & 170,288 & 9.86\% \\
\hline
4 & 65,727 & 3.80\% \\
\hline
5 & 18,616 & 1.07\% \\
\hline
6 \& more & 10,897 & 0.63\% \\
\hline
total & 1,726,694 & 100\% \\
\end{tabular}
\end{center}
\end{table}

Finally, it appears that 543,005 of the keyphrases, which constitute 31.44\% of all keyphrases, are not present in the body of the news articles. This number shows why we need keyphrase generation as well as extraction. As discussed before, in real-world instances, there are topics that are not explicitly mentioned in the text and spoken of implicitly.

\begin{table}
\caption{Number of Absent and Present Keyphrases}
\label{Table:absent_present}
\begin{center}
\begin{tabular}{c|cc}
& \# of keyphrases & \% of total \\
\hline
\hline
present & 1,183,689 & 68.55\% \\
\hline
absent & 543,005 & 31.44\% \\
\hline
total & 1,726,694 & 100\% \\
\end{tabular}
\end{center}
\end{table}

After shuffling, we divided the dataset into three subsets, 345,645 samples for training, 25,000 for validation, and 25,000 for test. The dataset can be found in this project's Github repository.

\subsection{Implementation Details}
We used dynamic word embeddings for the 100,000 most frequent words in the training set with embedding size of 150. Number of the units in the GRU cell for both the encoder and the decoder is 256. As for the beam search decoding hyperparameters, we set the beam width to 50 and beam depth to 5. Finally we used batch size of 32, Adam Optimizer with learning rate of $1e-3$ and gradient clipping by 0.1. We also used dropout of 0.5 as regularization on the input embeddings and the output of the encoder.

\subsection{Baseline Models}
Our baseline models constitute of unsupervised and supervised methods. As for the unsupervised models, we used graph-based models SingleRank \cite{wan2008single}, TopicRank \cite{bougouin2013topicrank}, and MultipartiteRank \cite{bougouin2013topicrank}. Unsupervised statistical methods include phrase-level TFIDF \cite{kim2013automatic}, KPMiner \cite{el2009kp}, and YAKE \cite{campos2018text,campos2018yake}. We used KEA \cite{witten2005kea} as our supervised base-line method. For more information on the hyperparameters, settings and implementation of the base-line models see \cite{doost2018perkey}.

\subsection{Evaluation Metrics}
Following the convention, we use precision, recall, and their harmonic average, F\textsubscript{1}-score, to evaluate the models. As these metrics may be too strict, we also measure the performance of the keyphrases using ROUGE-1 and ROUGE-2 \cite{lin2004rouge}. Using ROUGE metrics may help with the instances where there is only one (usually optional) word difference between gold keyphrase and the predicted one.

\section{Experimental Results}
We evaluated the performance of each method with their $k$ best outputs, $k$ being 5 and 10, using precision, recall and F\textsubscript{1}-score. As expected, RNN with GRU cell outperformed all the other supervised and unsupervised (statistical and graph-based) methods by a great margin. Table V contains the results for performance of all the mathods on whole dataset, i.e. absent and present keyphrases. After RNN, the best performance belongs to statistical methods, supervised and unsupervised, and lastly graph-based methods. TFIDF, as the simplest method, shows a good performance, especially regarding the recall. The best performance in each column is in bold, the second underlined, and the third in italics. 

\begin{table}[h]
\caption{Empirical Results on Both Absent and Present Keyphrases}
\label{Table:absent_present}
\begin{center}
\begin{tabular}{c|ccc|ccc}
\hline
\hline
Method & P@5 & R@5 & F\textsubscript{1}@5 & P@10 & R@10 & F\textsubscript{1}@10\\
\hline
TFIDF & .1724 & \emph{.2060} & .1877 & .1216 & \emph{.2877} & .1710 \\
KPMiner & \underline{.1900} & \emph{.1948} & .1924 & \underline{.1632} & .2513 & \underline{.1979}\\
YAKE & .0726 & .0820 & .0770 & .0658 & .1481 & .0911 \\
\hline
S.Rank & .0532 & .0671 & .0594 & .0623 & .1533 & .0886 \\
T.Rank & .0986 & .1208 & .1086 & .0665 & .1583 & .0937 \\
M.Rank & .1093 & .1319 & .1196 & .0771 & .1835 & .1086 \\
\hline
KEA & \emph{.1837} & \underline{.2226} & \underline{.2013} & \emph{.1300} & \underline{.3115} & \emph{.1835} \\
\hline
RNN & \bf{.3126} & \bf{.3888} & \bf{.3466} & \bf{.1947} & \bf{.4732} & \bf{.2759} \\
\hline
\hline
\end{tabular}
\end{center}
\end{table}

However, it is not fair to evaluate extractive methods, such as KPMiner and SingleRank, on absent keyprhases, as they fail to generate any. In table VI, we compare the evaluation results only on the present keyphrases. Here, RNN obtained more than twice the F\textsubscript{1}-score of the second-best approach, KEA, when $k$ is 5, and KPMiner, when $k$ is 10. RNN succeeded to obtain higher precision, compating to table V, which is the main reason for its exceptional F\textsubscript{1}-score, 44.69\%.

\begin{table}[h]
\caption{Empirical Results on Present Keyphrases}
\label{Table:absents_present}
\begin{tabular}{c|ccc|ccc}
\hline
\hline
Method & P@5 & R@5 & F\textsubscript{1}@5 & P@10 & R@10 & F\textsubscript{1}@10\\
\hline
TFIDF & .1726 & \emph{.2713} & .2110 & .1218 & \emph{.3792} & .1843 \\
KPMiner & \underline{.1902} & .2523 & \emph{.2169} & \underline{.1634} & .3240 & \underline{.2173} \\
YAKE & .0733 & .1089 & .0876 & .0664 & .1974 & .0994 \\
\hline
S.Rank & .0533 & .0996 & .0694 & .0624 & .2180 & .0970 \\
T.Rank & .0987 & .1634 & .1230 & .0665 & .2136 & .1015  \\
M.Rank & .1093 & .1761 & .1349 & .0771 & .2468 & .1176 \\
\hline
KEA & \emph{.1839} & \underline{.2937} & \underline{.2262} & \emph{.1301} & \underline{.4124} & \emph{.1979} \\
\hline
RNN & \bf{.5249} & \bf{.4794} & \bf{.5012} & \bf{.4621} & \bf{.5588} & \bf{.5059} \\
\hline
\hline
\end{tabular}
\end{table}

Table VII belongs to the results on the absent keyphrases. As other methods cannot generate absent keyphrases, we have only the results of RNN.

\begin{table}[b]
\caption{Empirical Results on Absent Keyphrases}
\label{Table:absents_present}
\begin{center}
\begin{tabular}{c|c}
\hline
\hline
P@5 & .1786 \\
R@5 & .2953 \\
F\textsubscript{1}@5 & .2226 \\
\hline
P@10 & .0960 \\
R@10 & .3697 \\
F\textsubscript{1}@10 & .1524 \\
\hline
P@20 & .0505 \\ 
R@20 & .4360 \\
F\textsubscript{1}@20 & .0905 \\
\hline
P@50 & .0219 \\
R@50 & .5155 \\
F\textsubscript{1}@50 & .0421 \\
\hline
\hline
\end{tabular}
\end{center}
\end{table}

\begin{table*}[ht]
\caption{Results of the ROUGE-1 and ROUGE-2 Metrics on Absent and Present Keyphrases}
\label{Table:absents_present}
\centering
\begin{tabular}{c|ccc|ccc|ccc|ccc}
\hline
\hline
& & & \multicolumn{2}{c}{ROUGE-1} & & & & & \multicolumn{2}{c}{ROUGE-2} & & \\
\hline
Method & P@5 & R@5 & F\textsubscript{1}@5 & P@10 & R@10 & F\textsubscript{1}@10 & P@5 & R@5 & F\textsubscript{1}@5 & P@10 & R@10 & F\textsubscript{1}@10 \\
\hline
TFIDF & .3634 & .2791 & \emph{.2983} & .2702 & .3824 & .2982 & .0551 & .0478 & .0476 & .0417 & .0784 & .0508 \\
KPMiner & \underline{.3989} & .2606 & .2912 & \underline{.3530} & .3243 & \emph{.3000} & \emph{.0552} & .0437 & .0447 & \emph{.0474} & .0631 & .0479\\
YAKE & .1899 & .2181 & .1869 & .1739 & .3368 & .2126 & .0331 & .0435 & .0340 & .0269 & .0695 & .0358 \\
\hline
S.Rank & .2040 & \underline{.3248} & .2359 & .1587 & \underline{.4245} & .2200 & .0498 & \underline{.0996} & \underline{.0619} & .0361 & \underline{.1304} & \emph{.0535} \\
T.Rank & .2325 & \emph{.3100} & .2503 & .1631 & \emph{.4042} & .2214  & .0493 & \emph{.0717} & .0541 & .0334 & .0920 & .0460 \\
M.Rank & .2197 & .2444 & .2175 & .1606 & .3519 & .2094 & .0447 & .0531 & .0448 & .0321 & .0794 & .0428 \\
\hline
KEA & \emph{.3814} & .2939 & \underline{.3139} & \emph{.2930} & .3996 & \underline{.3184} & \underline{.0646} & .0588 & \emph{.0572} & \underline{.0490} & \emph{.0964} & \underline{.0605} \\
\hline
RNN & \bf{.4443} & \bf{.4485} & \bf{.4245} & \bf{.3058} & \bf{.5544} & \bf{.3765} & \bf{.1957} & \bf{.2288} & \bf{.1995} & \bf{.1226} & \bf{.2800} & \bf{.1618} \\
\hline
\hline
\end{tabular}
\end{table*}

In table VIII you can see the results of the ROUGE-1 and ROUGE-2 metrics on the predictions. We measured the performance of the model using these metrics as suggested in \cite{hasan2014automatic}. The metrics may appear to be less strict as they show 8 percent more F\textsubscript{1}-score, but the precision drops when $k$ is 10, as the reference keyphrases decrease, and so does the F\textsubscript{1}-score, consequently. This shows that ROUGE metric may not be a better choice to measure the performance of the keyphrase extraction and generation models.

\section{Discussion}
We saw that the recurrent neural network with gated recurrent units could outperform other methods by a huge margin, especially in predicting present keyphrases. One may argue that despite the high performance of such algorithms, they require huge amount of training data. In this paper, we showed that we can overcome the barrier, having access to the vastest database of all times, internet. However, not all the supervised methods can produce such results. KEA, having access to the same training data as RNN, could not beat KPMiner, as it uses only two features, hence not powerful enough to do so.

Another problem we addressed in this paper was measuring the performance of the models. As discussed in other papers on keyphrase extraction and generation, the current performance measure is not perfectly suitable for the task, as it is too strict, meaning that it does not tolerate even a single word difference with the reference keyphrases, let alone understanding the different ways of expressing the same concept. The reference keyphrases themselves are another matter. Choosing them is quite subjective, in a way that different annotators most probably come up with different keyphrases, not even agreeing on the number of them. As an example to illustrate this, in table IX, you see the predicted keyphrases of a random news article and its reference keyphrases annotated by the author of it. For these predictions, the precision will be $0.2$ and recall $0.5$, you can see that they are of more quality however.

\begin{table}[ht]
\caption{An Example of Network's Predictions and the Reference Keyphrases of a Random News Article}
\label{Table:absents_presnent}
\centering
\begin{tabular}{c|c}
Predictions@10 &
Kordest\^{a}n-e Ar\^{a}q, tahavvol\^{a}t-e mantaqe, Ar\^{a}q,
\\
&
tahavvol\^{a}t-e Kordest\^{a}n-e Ar\^{a}q, Torkiye,
\\
&
tahavvol\^{a}t-e Ar\^{a}q, Ha\v{s}d-e-\v{s}\v{s}a'bi, D\^{a}'e\v{s},

\\
&
Rajab Tayyeb Ardo\u{g}\^{a}n, hameporsi
\\
&\\
References & 
Ben-Ali Ildirim, tahavvol\^{a}t-e Kordest\^{a}n-e Ar\^{a}q,
\\
&
Kordest\^{a}n-e Ar\^{a}q, Jah\^{a}ngiri
\end{tabular}
\end{table}

Most of these keyphrases are related to the main topics of the news article, expect for the last two. They are generated most probably because of some mentions of Iraq's `territorial integrity' and `security'. One interesting predicted keyphrase in this example is `Rajab Tayyeb Ardo\u{g}\^{a}n,' which is not mentioned in the article, except for an implicit mention of the title, `Turkey's president'.

We also measured the performance of the models using another suggested performance measure, ROUGE, which was not a better measurement, as discussed in section V.

\section{Conclusion and Future Work}
In this work, we compared the results of a couple of unsupervised and supervised methods on the tasks of keyphrase extraction and generation in Persian language, concluding that RNN with GRU cells can significantly outperform other methods in both tasks. We also measured the performance using ROUGE-1 and ROUGE-2 metrics, as they are less strict and may be more accurate and discuss their shortcomings.

As of the future work, we are going to address the task of multi-document keyphrase extraction and generation from Persian news articles.

\bigskip
\bibliographystyle{IEEEtran}
\bibliography{bibliography}

\end{document}